\pdfoutput=1

\documentclass[11pt]{article}

\usepackage{ACL2023}

\usepackage{times}
\usepackage{latexsym}

\usepackage{algorithm}
\usepackage{algpseudocode}
\usepackage{multirow}
\usepackage{booktabs}
\usepackage{xcolor}
\usepackage{graphicx}
\usepackage{amsmath}
\usepackage{amssymb}
\usepackage{mathtools}
\usepackage{bm}
\usepackage{arydshln}
\usepackage{tabularx}
\usepackage{hyperref}

\usepackage{microtype}

\newcommand{\ourmodel}{\textsc{Socratic CoT}}
\newcommand{\probP}{\mathbb{P}}

\makeatletter
\def\adl@drawiv#1#2#3{%
        \hskip.5\tabcolsep
        \xleaders#3{#2.5\@tempdimb #1{1}#2.5\@tempdimb}%
                #2\z@ plus1fil minus1fil\relax
        \hskip.5\tabcolsep}
\newcommand{\cdashlinelr}[1]{%
  \noalign{\vskip\aboverulesep
           \global\let\@dashdrawstore\adl@draw
           \global\let\adl@draw\adl@drawiv}
  \cdashline{#1}
  \noalign{\global\let\adl@draw\@dashdrawstore
           \vskip\belowrulesep}}
\makeatother

\usepackage{color}
\usepackage{soul}
\usepackage[textsize=scriptsize]{todonotes}

\title{
Distilling Reasoning Capabilities into Smaller Language Models
}

\author{%
  Kumar Shridhar $^{*}$  \quad Alessandro Stolfo \thanks{\quad  Equal contribution;} \quad Mrinmaya Sachan \\
  Department of Computer Science, ETH Z\"urich \\ 
  \texttt{\{shkumar, stolfoa\}@ethz.ch}}

\date{}

\begin{document}
\maketitle
\begin{abstract}

Step-by-step reasoning approaches like chain of thought (CoT) have proved to be very effective in inducing reasoning capabilities in large language models. 
However, the success of the CoT approach is fundamentally tied to the model size, and billion parameter-scale models are often needed to get CoT to work. 
In this paper, we propose a knowledge distillation approach that leverages the step-by-step CoT reasoning capabilities of larger models and distills these abilities into smaller models. 

In this work, we propose an alternative reasoning scheme, \ourmodel\ that learns a decomposition of the original problem into a sequence of subproblems and uses it to guide the intermediate reasoning steps. 
We use \ourmodel\ to train a combination of two small distilled models: a \textit{problem decomposer} and a \textit{subproblem solver}.
In practice, given a new problem, the two distilled models work in sync to decompose and solve complex problems.
On multiple reasoning datasets (GSM8K, StrategyQA, and SVAMP), our proposed distillation strategies boosts the performance of smaller models over 70\% compared to the baselines. 
Finally, we investigate when \ourmodel\ is an effective alternative to CoT, demonstrating cases where a much smaller model (GPT-2 large) can outperform a 10X larger model (GPT-3 6B). Our code is available \href{https://github.com/kumar-shridhar/Distiiling-LM}{here}.

\end{abstract}

\section{Introduction}
 
Large language models (LLMs) have demonstrated strong performance on a variety of reasoning tasks \cite[\textit{inter alia}]{brown2020language, hoffmann2022training, chowdhery2022palm}. 
One particularly interesting strategy for prompting these models is chain-of-thought (CoT), which has been shown to elicit reasoning abilities in LLMs by asking the model to incorporate intermediate reasoning steps while solving a problem \cite{nye2021show, wei2022chain, Wang2022SelfConsistency}.
However, CoT has been shown to work primarily on models with hundreds of billions of parameters  
\cite{wei2022chain, wei2022emergent} or those tuned to a wide range of tasks \cite{flan-t5, iyer2022opt}. 

\begin{figure}[h!]
    \centering
    \includegraphics[width=0.44\textwidth]{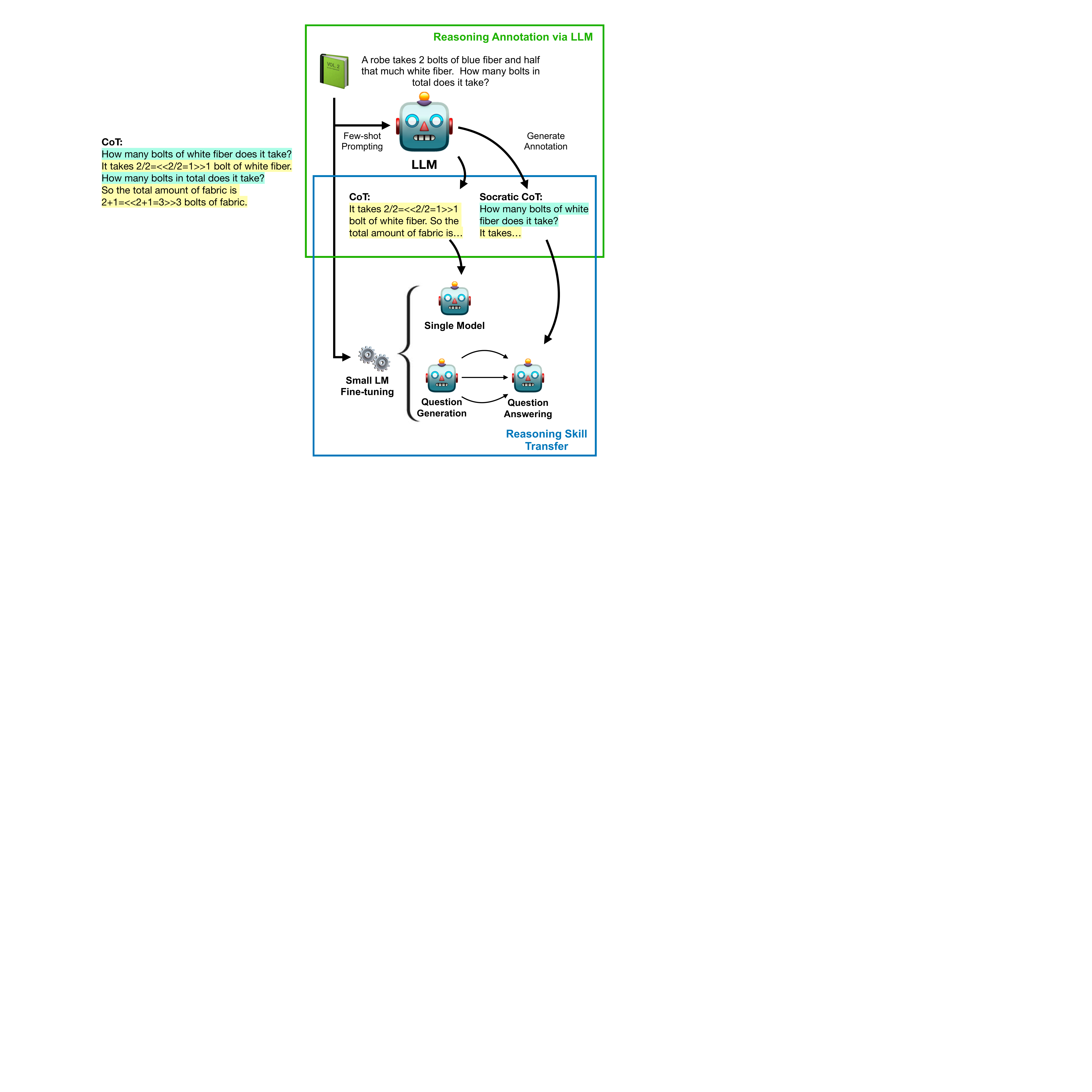}
    \caption{Illustration of the proposed framework. First, an LLM is prompted to decompose a multi-step problem providing annotation for the intermediate steps leading to the final solution. Then, the generated annotation is used to provide additional supervision when fine-tuning smaller models.
    }
    \label{fig:intro}
\end{figure}

Due to the significant computational resources or expensive API calls required to access CoT-capable LLMs, we ask whether it is possible to elicit such reasoning capabilities in smaller models.\footnote{Following \citet{li2022explanations}, we argue that \textit{small} and \textit{large} models are relative terms and context-dependent. We consider models with billions of parameters to be large, and models with millions of parameters to be small.}

Small-sized, non-fine-tuned language models are known to be poor reasoners \cite{stolfo2022causal}. Therefore, a possible approach to induce CoT-like reasoning abilities in smaller models would be fine-tuning them on step-by-step examples.

In our work, we propose a framework for leveraging the reasoning capabilities of LLMs to supervise the training of smaller models. This approach can be thought of as a form of \textit{knowledge distillation} \cite{hinton2015distilling}, where a larger teacher model transfers knowledge to a smaller student model.
However, unlike standard knowledge distillation, our method transfers the reasoning abilities of the teacher model only using its generated solutions as a proxy, i.e., we do not assume access to the teacher model parameters.
Our approach consists of prompting an LLM to produce step-by-step annotations leading to the answer for a set of problems. This annotation is then used as supervision to fine-tune the student model. A high-level illustration of the process is provided in Figure \ref{fig:intro}.

Within this framework, we study three different types of \textit{annotation structure} for supervising our distillation approach:
(i) We consider fine-tuning on the \textit{gold} step-by-step solution procedure for datasets where the step-by-step solutions are available. (ii) We study whether procedural supervision, coming from the chain of thought (CoT) of the teacher model can improve upon the baseline. (iii) We propose a third type of supervision structure, which we call \ourmodel. This approach relies on learning a semantic decomposition of the original problem into a sequence of subproblem-solution pairs using two models -- a) a question generator that learns to decompose the problem into a sequence of subproblems, and b) a question-answering model that solves the various generated subproblems (more details are in section \ref{socratic-cot}). This approach can be thought of as an extension of the typical chain of thought reasoning where, unlike CoT, the intermediate steps are now decomposed into subquestion-solution pairs; the subquestions guide the generation of intermediate steps that lead to the final answer to the problem.

We train distilled student models with various annotation structures mentioned above. Depending on the annotation available for the given data, we use the teacher model to generate either a CoT-like solution to a problem or, if the step-by-step annotation is available, a set of subquestions leading to the solution of the problem, or both (examples of different annotations are shown in Figure \ref{fig:socratic_cot}).

We perform our analyses on three multi-step reasoning datasets: GSM8K \cite{cobbe2021training}, StrategyQA \cite{geva2021strategyqa}, and SVAMP \cite{patel2021nlp}. We consider data with various types of annotation 
to cover a range of realistic data scenarios. Our results show that supervision by CoT-decomposed examples helps smaller models perform better, and subquestioning introduced by \ourmodel\ can provide further improvement. We observe performance gains of up to 40\% with LLM-generated step-by-step annotations -- this validates the effectiveness of our distillation framework (detailed analysis in Section \ref{results}).

\section{Related Work}
\paragraph{Decomposing Multi-Step Reasoning Tasks} Solving multi-step reasoning tasks like MWPs has been a popular area of research for the last couple of years \cite{kushman-etal-2014-learning, hosseini2014learning, roy-etal-2015-reasoning, 
amini2019mathqa, 
zhang2020graph,
shridhar2022automatic, opedal-et-al-2023}.  
However, the majority of the modern approaches for these problems are shifting towards using large language models, often relying on approaches involving prompting or in-context learning \citep{cobbe2021training,kojima2022large,wei2022chain,chowdhery2022palm, lewkowycz2022solving, srivastava2022beyond}.
One such prompting approach is the chain of thought prompting \cite{wei2022chain}, which prompts the language model to generate a series of %
intermediate steps that improve the reasoning capabilities in LLMs. \citet{Wang2022SelfConsistency} took another step forward and sampled multiple reasoning paths and selected the most relevant output using majority voting. \citet{huang2022large} used the most voted outputs to further fine-tune the model for better performance. \citet{kojima2022large} further improved the reasoning of LLM in a zero-shot manner by appending ``Let's think step by step'' to the prompt. In contrast, our work does not %
propose prompting solutions; instead, we explicitly guide the student model reasoning using sub-questions at each step. %
Most similar to our work is the work by \citet{Zhou2022LeasttoMostPE} which decomposes questions into sub-questions and asks the language model to solve each sub-question sequentially. However, this work is also restricted to prompting and only works with LLMs with billions of parameters. 

\paragraph{Knowledge Distillation} Our approach is reminiscent of knowledge distillation~\citep{ba2014deep,hinton2015distilling} in that we use a student network to mimic the large teacher language model. \citet{snell2022learning} demonstrated the usefulness of providing instruction that can help models achieve better reasoning skills. Similar to our hypothesis, \citet{ eisenstein2022honest} argued that question-answering systems should focus not only on the final answer, but also on the rationale that justifies their reasoning, to help them reason better. We go beyond this; in our work, in addition to the question-answering system, we also focus on what questions need to be asked at each step that can help to learn that reasoning step better. Finally, similar to our hypothesis of injecting reasoning capabilities into smaller models, \citet{li2022explanations} used CoT-like reasoning from LLMs to train smaller models on a joint task of generating the solution and explaining the generated solution. We, on the other hand, use the LLM to generate subquestions and solution pairs and use them together to inject reasoning capabilities into smaller models. 
\paragraph{Subquestioning as supervision} The idea of inquiring or asking information-seeking questions for discovery learning has been studied well in the past \cite{bruner1961act}. \citet{rao_answer-based_nodate} generated clarification questions based on Stack Exchange questions as supervision, \citet{klein2019learning} used a joint question answering model to ask questions from a given span of text and later answer them, and \cite{rajani_explain_2019, shwartz-etal-2020-unsupervised} asked questions to improve common sense QA models.  In contrast, our work focuses on multistep reasoning tasks where intermediate clarifying questions and reasoning steps may not always be available and may need to be extracted from a teacher model.

\section{Methodology}

The setting we consider consists of a data set $\mathcal{D}$, where each problem $P_i$ is accompanied by a final answer $a_i$ that can be reached by several steps of reasoning.
The task of solving the problem using a model $\psi$ is to predict an answer $\hat{a} = \psi(P)$ such that $\hat{a} = a$.
We consider different data scenarios where intermediate annotations of the solution may be available in different forms (e.g., step-by-step, as a semantic decomposition by subquestions) or may not be present. Depending on the availability of annotations, we propose different approaches to augment the training of a small model on $\mathcal{D}$ by using LLMs.

\begin{figure}[t]
    \centering
    \includegraphics[width=0.4\textwidth]{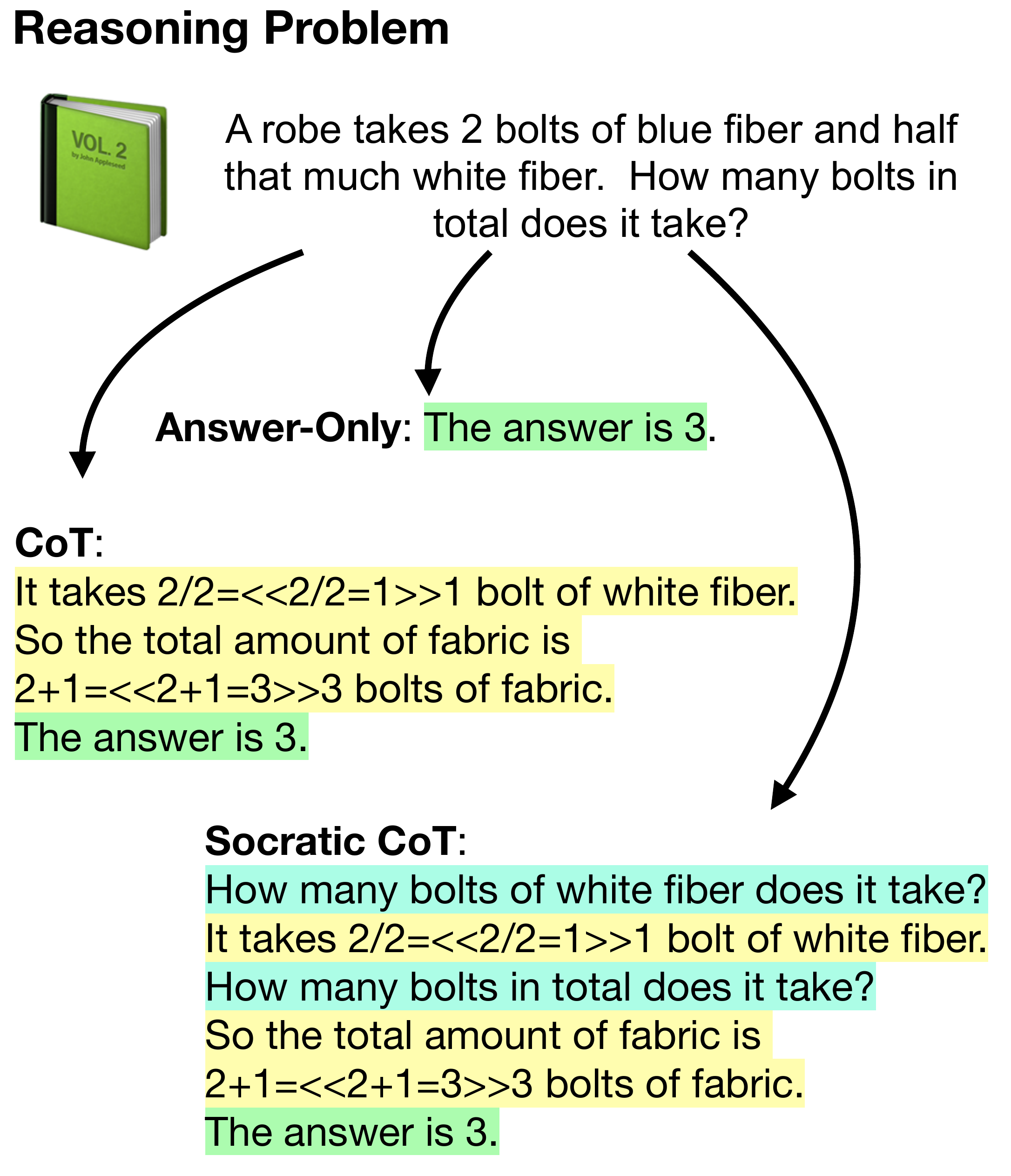}
    \caption{Illustration of the three different kinds of annotation structure. Our proposed approach, \ourmodel, augments the typical chain-of-thought step-by-step solution with subquestioning. }
    \label{fig:socratic_cot}

\end{figure}

\subsection{Distilling step-by-step reasoning via CoT}
\label{sec:cot}

A data set may present an annotation that contains intermediate reasoning steps that lead to the answer $a_i$ (i.e., a chain-of-thought annotation). This intermediate annotation can be used directly to fine-tune a small model. However, in cases where such step-by-step information is not available, we use a LLM to generate the reasoning steps that might improve the performance of the small model.

To achieve this, we consider a small subset of the dataset $\mathcal{D}$ and decompose each problem $P_i$ into $n_i$ intermediate reasoning steps.
We construct these intermediate reasoning steps manually, since we only need a few examples as prompts (examples are provided in Appendix Table \ref{table:prompts}).

For each remaining problem $P \in \mathcal{D}$, we then prompt a large language model $\mathcal{M}$ to generate the intermediate reasoning steps. 
We make sure that the chain of reasoning steps is meaningful by checking whether the last solution matches the ground truth answer, i.e. whether $a_i^{(n_i)} = a_i$, where $a_i^{(n_i)}$ represents the answer corresponding to the last reasoning step. If this is not the case, we discard the problem and sample a new chain by prompting the model again (for a maximum of 3 times).
In this way, we obtain an augmented dataset $\mathcal{D}^*$ in which a subset of problems is paired with a sequence of reasoning steps leading to the correct result.
Finally, we can distill the reasoning capabilities into smaller models by fine-tuning them with the generated intermediate steps.

\begin{figure*}[h]
    \includegraphics[width=\textwidth]{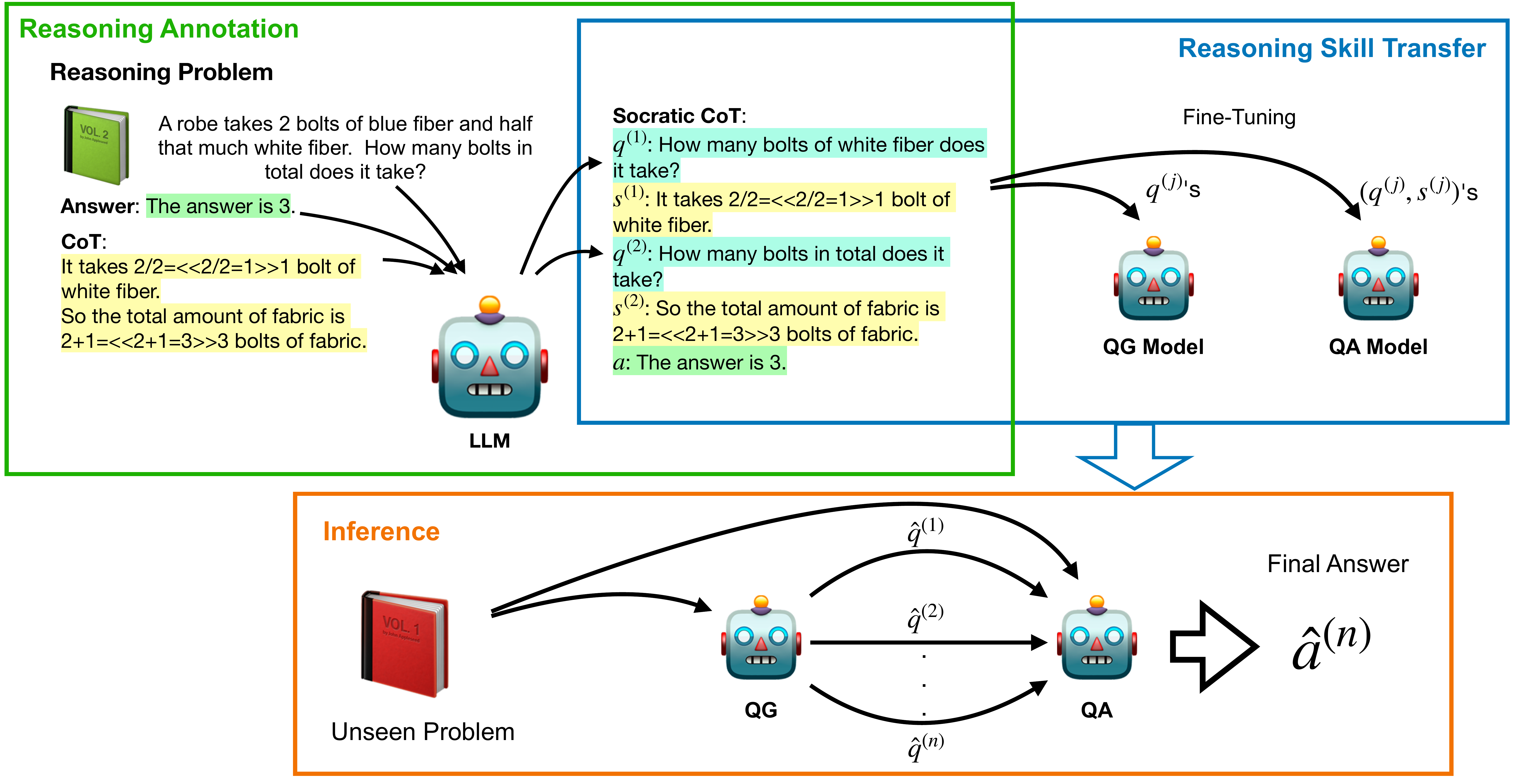}
    \caption{Detailed explanation of our framework. First, a LLM is prompted to decompose the input problem $P$ into a series of subquestion-solution pairs ($q_i^{(j)}, s_i^{(j)}$, $j \in \{1,\dots,n_i\}$) with an answer at each step $a_i^{(j)}$. The generated subquestions-solutions are used to train two student models: a) the QG model which learns to mimic sub questioning capability of the LLM and b) the QA model, which learns to solve each subquestion. At the bottom, the inference process is depicted for an unseen problem and no LLM is involved. The QG model breaks the unseen problem into simpler subquestions and the QA model solves each one of them eventually leading to the final answer $a_i^{(n_i)}$.}
    \label{fig:main:large}
\end{figure*}

\subsection{Distilling step-by-step reasoning through \ourmodel}
\label{socratic-cot}
In this section, we describe how CoT can be enhanced through subquestioning. An illustration of our approach is shown in Figure \ref{fig:main:large}.
\subsubsection{Extracting the Reasoning Capability from the Teacher}

In Section \ref{sec:cot}, we detailed how an LLM can be used to generate the intermediate annotation of a problem $P_i$ as a chain of steps leading to the answer $a_i$.
We now extend this procedure to include a subquestion at each step of the solution.
Following a similar procedure as described in Section \ref{sec:cot}, we prompt the LLM with few exemplars of problems decomposed as a set of intermediate subquestion-solution pairs (the prompts are reported in Appendix Table \ref{table:prompts}).
This way, we obtain an intermediate annotation that includes subquestioning. In particular, each of the $n_i$ steps constituting the overall solution is a subquestion-solution pair, denoted $q_i^{(j)}, s_i^{(j)}$, $j \in \{1,\dots,n_i\}$ (an example is shown in Figure \ref{fig:socratic_cot}). We refer to the ordered list of subquestion-solution pairs for problem $P_i$ as $(q_i^{(1)}, s_i^{(1)}), \dots, (q_i^{(n_i)},s_i^{(n_i)})$.

\subsubsection{Transferring the Reasoning Capability into the Student}
We present two strategies to distill the reasoning annotation provided by the LLM into smaller models. 

In the first strategy, a single \textit{unified} student is trained to generate the subquestion-solution pairs simultaneously, while in the second strategy, the question generation and question-answering tasks are assigned to two separate models. We call this second strategy \textit{iterative} because the question-answering model is trained to solve each subquestion iteratively.

\paragraph{Unified.}
Using the problems in $\mathcal{D}$ that contain the chain of intermediate questions and solutions, we train a \textit{unified} student model $\mathcal{M}_{uni}$ that learns to generate the sequence of subquestion-solution pairs $\{(q^{(1)},s^{(1)}), (q^{(2)}, s^{(2)}), \dots \}$ that lead to the solution of a given problem. We use a pre-trained transformer-based model \cite{vaswani2017attention} and train it on the chain of subquestion-solution pairs for each problem $P$. 
Given a step $j$ of problem $P$ (i.e., the concatenation of $q^{(j)}$ and $s^{(j)}$) consisting of a sequence of $m_j$ tokens $\{x_j^{(1)}, \dots, x_j^{(m_j)}\}$, we use a typical auto-regressive language modeling loss, $\mathcal{L}$:

\begin{align}
\label{eq:loss_uni}
    & \mathcal{L}_j(P) = 
    & - \sum_{k=1}^{m_j} \log \probP_{{uni}}\ (x_j^{(k)}|x_j^{:(k-1)}, P)
\end{align}
where $ \probP_{{uni}}(x | c)$ is the probability assigned by $\mathcal{M}_{uni}$ to token $x$ given context $c$, and $x^{:(y)}$ indicates the sequence $\{x^{(1)}, \dots, x^{(y)}\}$. The loss $\mathcal{L}_j$ is computed for each problem $P_i$ and for each pair $(q^{(j)},s^{(j)})$ %
leading to the final answer $a_i$. 

\begin{table*}[t]
\centering
\small
\begin{tabularx}{\textwidth}{X X}
    \toprule
     \multicolumn{2}{c}{\textbf{Unified}}  \\
     \midrule
     Input: & Output: \\
      A robe takes 2 bolts of blue fiber and half that much white fiber. How many bolts in total does it take? &  How many bolts of white fiber does it take? It takes 2/2 = $<<$2/2=1$>>$ 1 bolt of white fiber. How many bolts in total does it take? So the total amount of fabric is 2+1 = $<<$2+1=3$>>$ 3 bolts of fabric. The answer is 3.\\
    \midrule
     \multicolumn{2}{c}{\textbf{Iterative}}  \\
     \midrule
    \multicolumn{2}{c}{Iteration 1} \\
    Input: & Output:\\
    A robe takes 2 bolts of blue fiber and half that much white fiber. How many bolts in total does it take? & \textbf{QG}: How many bolts of white fiber does it take? \newline \textbf{QA}: It takes 2/2 = $<<$2/2=1$>>$ 1 bolt of white fiber. \\ \cdashlinelr{1-2}
      \multicolumn{2}{c}{Iteration 2} \\
      Input: & Output:\\
     A robe takes 2 bolts of blue fiber and half that much white fiber. How many bolts in total does it take? How many bolts of white fiber does it take?  It takes 2/2 = $<<$2/2=1$>>$ 1 bolt of white fiber. &
      \textbf{QG}: How many bolts in total does it take? \newline \textbf{QA}: So the total amount of fabric is 2+1 = $<<$2+1=3$>>$ 3 bolts of fabric. The answer is 3.\\
     
     \bottomrule
\end{tabularx}
    \caption{Example demonstraing the input-output format for unified vs iterative setup. QG represents the question generation model and QA is the question answerer mdoel. Note that QA model uses the QG output to answer it as shown in Figure \ref{fig:main:large}. }
    \label{table:unifiedvsiter}
\end{table*}

\paragraph{Iterative.}
The \textit{iterative} version of the student separates the tasks of generating the subquestions and providing an intermediate answer to each subquestion into two distinct models: a question generation (QG) model and a question answering (QA) model.
Both the QG and QA models are implemented using a Transformer-based language model \cite{vaswani2017attention}. In particular, the QA model $\mathcal{M}_{qa}$ is iteratively trained to answer the teacher-generated sub-questions. The learning objective is computed at the token level for each intermediate solution:

\begin{equation}
\resizebox{.95\linewidth}{!}{%
$\begin{aligned}
    \mathcal{L}(P, s^{(j)}) =
     - \sum_{k=1}^{l_j} \log \probP_{\mathcal{QA}}\ (y_j^{(k)}|y_j^{:(k-1)}, q^{:(j)}, s^{:(j-1)}, P)
     \nonumber
\end{aligned}$}
\end{equation}

where $l_j$ and the $y_j$'s represent, respectively, the length and the tokens of the intermediate solution $s^{(j)}$. $s^{:(j-1)}$ consists of the previous solution generated by the QA model iteratively in the past iterations.

Similarly, the QG model is trained to acquire the ability of the teacher model to decompose the problem's main question into a series of sub-steps, each of which corresponds to a subquestion. The loss for this model is analogous to Equation \ref{eq:loss_uni}, with the only difference being that the intermediate solutions are not considered for the QG model.
During training, the previous intermediate solutions generated by the QA model are replaced with the teacher-generated solutions using teacher forcing \cite{cho2014learning}. However, the intermediate solutions generated by the model are used at inference time.

\subsection{Inference-time Predictions}

Given an unseen problem $P$, the unified student model can directly predict a solution as a sequence of subquestions and answers.
In the iterative approach, we first generate the subquestions conditioning the generation of the QG model on $P$.
After these questions are generated, they are provided to the QA model one by one, decoding the intermediate solution $\hat{s}^{(j)}$ at step $j$ token by token according to the model's probability distribution over its vocabulary:
\begin{align}
    \probP_{\mathcal{QA}}\ (y_j^{(k)}|y_j^{:(k-1)}, \hat{q}^{:(j)}, \hat{s}^{:(j-1)}, P),
\end{align}
where $y_j^{(k)}$ is the $k$-th token being decoded in greedy fashion.

After the last solution $\hat{s}^{(n)}$ has been generated, the numerical prediction $\hat{a}^{(n)}$ is parsed from the text using simple heuristics. 

\section{Empirical Analysis}

\subsection{Datasets}

We study how smaller models can learn to reason better on three multi-step reasoning datasets: GSM8K \cite{cobbe2021training}, StrategyQA \cite{geva2021strategyqa}, and SVAMP \cite{patel2021nlp}. GSM8K consists of 8.5K grade school math word problems, each requiring 2 to 8 steps of reasoning to solve. The solutions primarily involve a sequence of elementary calculations using basic arithmetic operations ($+$, $-$, $\times$, $\div$). The dataset is divided into 7.5K training problems and 1K test problems. 
To evaluate the model on SVAMP, we train the model on 761 multi-step math word problems taken from the ASDiv \cite{miao-etal-2020-diverse} training set and evaluate it on 237 multi-step SVAMP problems. 
For StrategyQA, the test set with facts is not available, so we split the data into 80\% training, 10\% as validation data, and the last 10\% as test data. We do not shuffle the data to maintain reproducibility.

\subsection{Experimental Setup}

We use three kinds of annotation, corresponding to the three datasets that we consider.
\paragraph{Step-by-step solution}: The GSM8K dataset falls into this category and includes a Socratic version where intermediate subquestion-solution pairs are provided for each MWP.
While the intermediate step-by-step solutions were manually annotated, the authors report that the subquestions were generated by prompting GPT-3.
We reproduced a subset of these subquestions using a GPT-3 model with prompts, and we observed a high similarity between the questions provided and the ones generated by us (BERT $F_1$ score of 95\%). For \ourmodel, we thus use the subquestioning annotation already provided.  
\paragraph{Supporting facts}: We study the StrategyQA dataset, which falls in this category. Strategy QA consists of a factual question with binary True/False as the final answer. Additional supporting facts and decomposed questions are provided. However, the set of facts and the decomposed questions provided with a given question are not always aligned (i.e., a fact is not necessarily the answer to one subquestion). Therefore, having a setup similar to the one for GSM8K is not possible. We thus consider two versions of the data. One in which the supporting facts are used as CoT and the corresponding questions are generated by prompting a GPT-3 model, and a second in which we take the provided questions and generate the facts (this time aligned with the questions) using GPT-3. 
\paragraph{Final answers only}: AsDiv/SVAMP falls in this category and for training, we use GPT-3 to generate both intermediate subquestions and solutions. Intermediate solutions are used as CoT and the generated subquestion-solution pairs for \ourmodel.

\subsection{Implementation Details}
We use GPT-2 variants \cite{radford2019language} as student models. GPT-3 175B \cite{brown2020language} served as the teacher model for decomposing complex problems into a series of simpler substeps (we report the prompts used in Appendix Table \ref{table:prompts}). 

All models were trained using the Huggingface library \cite{wolf2019huggingface} on an NVIDIA Tesla A100 GPU with 40 GB of memory. Each experiment was run for the same number of iterations to ensure fairness with periodic evaluation over the validation set. Teacher forcing was used during training to replace the generated responses with ground truth answers from the training dataset. 
\paragraph{Evaluation Metric.}
To evaluate the question-answering performance on the GSM8K, SVAMP, and StrategyQA datasets, we compute the accuracy based on the final answer provided by the student model.

\begin{table*} [t]
\centering
\resizebox{\textwidth}{!}{
\begin{tabular}{c | c  c c c | c c c c}
    \toprule 
     &  &  &  & & & \multicolumn{2}{c}{Iterative} & Unified\\
    \bf{Dataset} &  \bf{Model} & \bf{Answer Only} & \bf{GT Steps} & \bf{GT Facts} & \bf{CoT} & \bf{Soc}$_{CoT}$ & \bf{Soc}$_{GT}$ & \bf{Soc}$_{CoT}$\\
    \midrule
    & Small (124M)  &  1.45 & 5.05 & - & 4.70 & 5.98 & \bf 6.44 ($\uparrow 20\%$) & 5.10 \\ 
    \bf GSM8K & Medium (355M) &  2.90  & 7.88 & - &  7.10 & 11.57 & \bf{12.74} ($\uparrow 38\%$)  & 7.90  \\
    & Large (774M) &  4.62 & 14.10 & -& 12.85 & 17.89 & \bf 21.08 ($\uparrow 33\%$) & 13.25 \\ \cdashlinelr{2-9}
    & GPT-3 (6B) &  - & 21.00 & - & - & - & - & - \\
    \midrule
     & Medium (355M)  &  54.10 & - & 52.02 & 55.01 & 52.05 & \bf{60.31} ($\uparrow 13\%$)  & 52.05   \\
    \bf StrategyQA & Large (774M) &  61.10 & - & 62.80 & 55.90 & 61.32 & \bf{66.40} ($\uparrow 5\%$)  & 59. 45\\
    & XL (1.5B)  &  60.51 & - & \bf 66.30 & 58.07 & 62.30 & 63.56 ($\downarrow 4\%$) & 62.05   \\
    \midrule
     & Small (124M)  &   2.15 & - & - & 5.35 & \bf{6.79}  & -  & 5.82    \\
    \bf SVAMP & Medium (355M) & 4.80 & - & -& 17.30 & \bf{18.99} & - & 17.62  \\
    & Large (774M) &  7.40 & - & - &  \bf 23.60  & 18.14 & - & 17.45  \\
    
    \bottomrule
\end{tabular}
}
\caption{Accuracy comparison (in \%) on the three considered datasets. We consider three human-annotated baselines: final answers only (Answer Only), ground-truth step-by-step solution (GT Steps), and supporting facts (GT Facts). We compare the different supervision strategies for fine-tuning the small models: \textbf{CoT} represents the case where the chain of intermediate reasoning steps is generated by GPT-3, \textbf{Soc}$_{CoT}$ represents the case where both the chain of intermediate solutions and the subquestions are generated by LLM and used to fine-tune small models. \textbf{Soc}$_{GT}$ represents the case where GT solutions/facts are used when prompting GPT-3 to generate the subquestions. Iterative and Unified represent the two \textbf{Soc}$_{CoT}$ strategies described above. All models are GPT-2 versions and their size is reported within parentheses. All experiments were run at least 3 times and the average is reported. GPT-3 6B results are taken from \citet{cobbe2021training}.}
\label{Acc:all_models}
\end{table*}

\section{Results and Discussion}
\label{results}
\paragraph{Can our framework improve the reasoning capabilities of smaller models?}

Table \ref{Acc:all_models} demonstrates that leveraging LLMs reasoning capabilities using our framework can improve the reasoning results for all dataset types. 

\begin{figure}[t]
    \centering
    \includegraphics[width=0.47\textwidth]{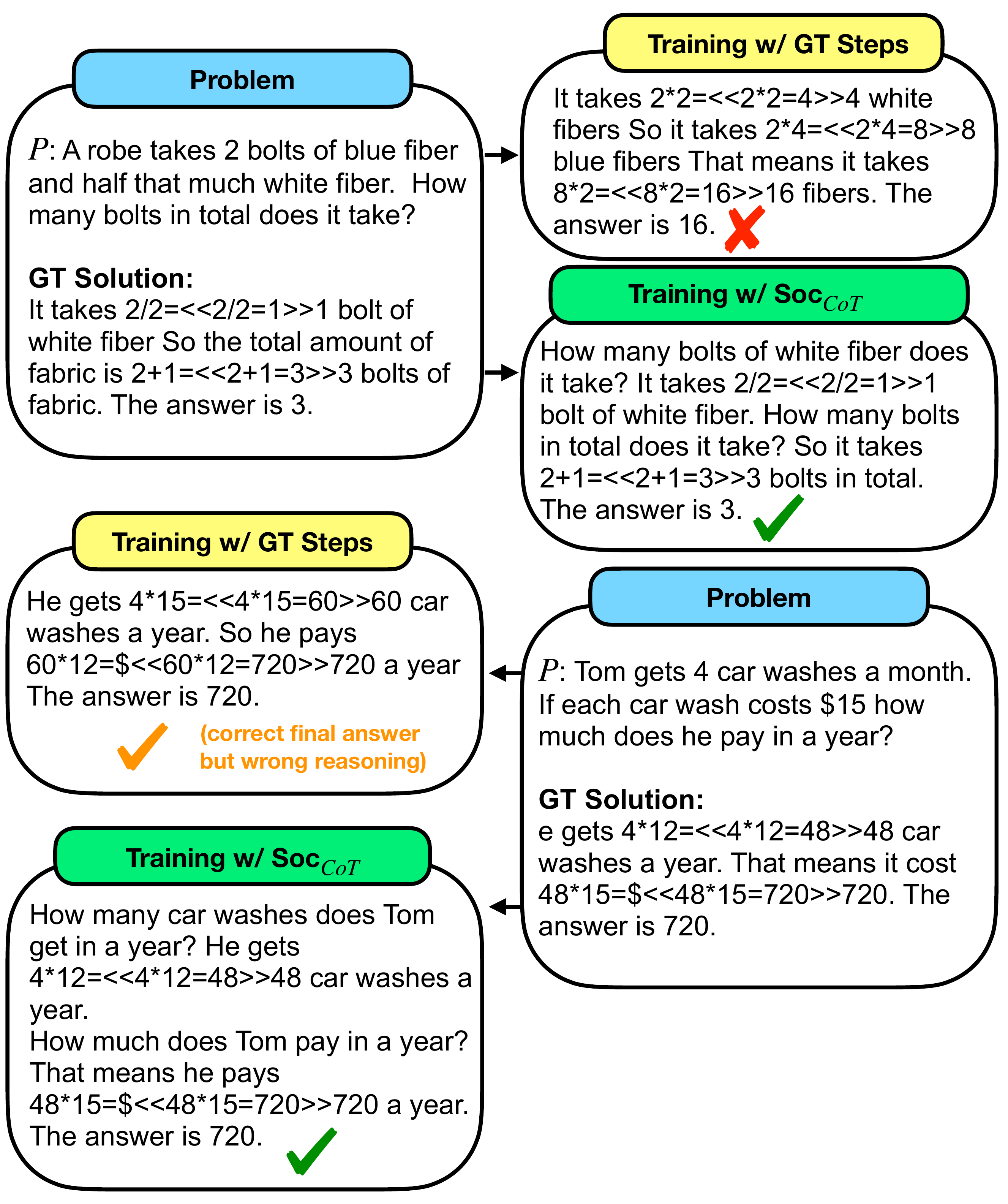}
    \caption{Example of predictions generated by a GPT-2 Large model fine-tuned with GT steps and \ourmodel\ on GSM8K dataset.}
    \label{fig:examples}
\end{figure}

\paragraph{Step-by-Step Solution.} When human-annotated step-by-step solutions are available, training smaller models with LLM-generated CoT is not advantageous, as shown on GSM8K. This is to be expected since the annotation generated by an LLM is likely to be noisier and of lower quality than human-annotated data.
However, the ground-truth step-by-step annotation can be leveraged to prompt an LLM to generate subquestions for the \ourmodel\ approach, giving a performance boost of up to 38\% when the LLM-generated subquestions are used at inference time.
When the subquestions are learned by the QG model (Iterative \textbf{Soc}$_{CoT}$), the accuracy of the student model decreases slightly but still improves over the step-by-step annotation without subquestions (17.89 vs. 14.10). Figure \ref{fig:examples} shows a comparison of predictions generated by \textbf{Soc}$_{CoT}$ models and a model trained on the GT step-by-step annotation.
Unified \ourmodel\ performs similarly to training with the step-wise ground-truth annotation. We additionally include the score produced by GTP-3 6B to show that training with \ourmodel\ can help a small model (GPT-2 large with 774M parameters) perform as well as a nearly 10x larger model fine-tuned with human annotated data.

\begin{figure}[t]
    \includegraphics[width=0.45\textwidth]{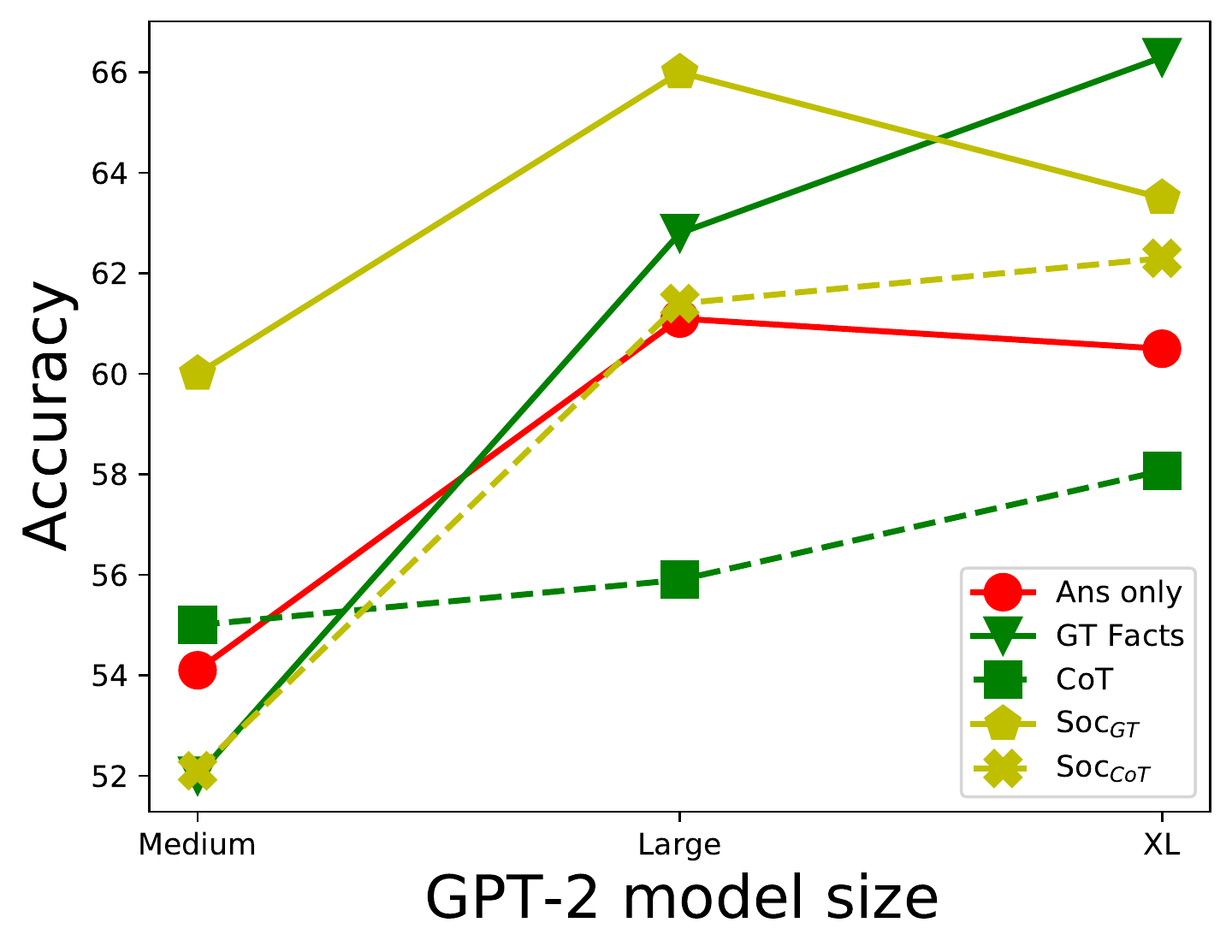}
    \caption{Accuracy comparison for different supervision strategies on StrategyQA. The baseline method consists of fine-tuning on final answers only (Ans only), and it is compared to fine-tuning with: ground-truth supporting facts (GT Facts), GPT-3-generated supporting facts (CoT), ground-truth supporting facts with GPT-3-generated subquestions (\textbf{Soc}$_{CoT}$), and LLM-generated facts with human-annotated subquestions (\textbf{Soc}$_{GT}$).
    }
    \label{fig:strategy}
\end{figure}

\paragraph{Supporting facts.} On StrategyQA, we observe that the inclusion of ground-truth supporting facts in the fine-tuning procedure improves the performance of the small models.
However, surprisingly, when the supporting facts are generated by GPT-3, their inclusion actually hurts performance (58.07 vs 60.51 for GPT-2 Large).
We hypothesize that this is likely due to the imperfect factual knowledge provided by the LLM, which mars the quality of the supervision.
We have observed that the GT supporting facts provided often do not represent a logical sequence of propositions leading to the final answer. This is likely the reason why decomposing the problem through subquestions based on such facts actually harms accuracy (see \textbf{Soc}$_{CoT}$ column in Table \ref{Acc:all_models}).
Instead, using the provided subquestions and using an LLM to generate the answers (representing coherent facts leading to the final answer) proves to be an effective strategy (60.31 vs. 52.02 for GPT-2 Medium).
A more detailed comparison between our proposed approaches is presented in Figure \ref{fig:strategy}.
However, GPT-2 XL models perform well when trained on facts as unlike smaller models, larger models can encode more facts at once in their parameters, which assists in answering a factual question.

\paragraph{Answers only.} On the SVAMP dataset, which includes only final answers and no intermediate annotation,
LLMs can be used to generate both the intermediate steps and the subquestions. Both the consideration of intermediate solutions without subquestions (\textbf{CoT}) and the consideration of intermediate solutions with subquestions (\textbf{Soc}$_{CoT}$) lead to an improvement in performance.
The trend here is similar to what was observed for StrategyQA, with \ourmodel\ being more effective for the two smaller models but falling back to \textbf{CoT} for the larger model. 

\begin{table} [t]
\centering
\small
\begin{tabular}{c | c c  }
    \toprule
    { \bf Models } & \bf Methodology  & {\bf Accuracy } \\
    \midrule
    GPT-3 (1-shot)  & CoT   & 27.5      \\\cdashlinelr{2-3}
    (175B) & Sub-ques   & \textbf{47.1} ($\uparrow 41\%$)    \\ 
    \bottomrule
\end{tabular}

\caption{Accuracy comparison (in \%) of using CoT vs \ourmodel\ (Sub-ques) on the GSM8K dataset for GPT-3 model with prompting.}
\label{Acc:GPT-3}
\end{table}

\paragraph{Can \ourmodel\ be used as a prompting strategy?}
We experimented with \ourmodel\ as a prompting strategy. First, we prompted GPT-3 (175B) to decompose the main problem into simpler steps by formulating subquestions. Then, GPT-3 is used again to solve the sequence of subproblems in a single-shot setting with a problem decomposed into intermediate subquestions and solutions included in the prompt. The introduction of subquestioning boosts accuracy by over 40\% compared to standard CoT prompting (Table \ref{Acc:GPT-3}).
Other work (e.g., \citealt{wei2022chain}) has used a larger number of exemplars in the few-shot prompt, achieving higher overall accuracy. We limited our experiments to single-shot prompts due to budget constraints.

\section{Ablation Studies}
In this Section, we describe additional analyses regarding specific components of the framework we propose, as well as negative results that we obtained with alternative strategies.

\paragraph{How good are the sub-questioning capabilities of a smaller model?}
We investigate in more detail the ability of a small model to decompose a problem by generating meaningful subquestions. We fine-tuned GPT-2 Large on the GPT-3 generated subquestions provided in the GSM8K dataset. We then evaluated the quality of the generated questions in terms of BLEU score \cite{sacre-bleu}, BERT F$_1$ score \cite{zhang2019bertscore}, and by measuring for how many problems the number of questions generated by GPT-2 (\#Q) matches the number of GPT-3 annotated questions for a given problem.

\begin{figure}[t]
    \includegraphics[width=0.5\textwidth]{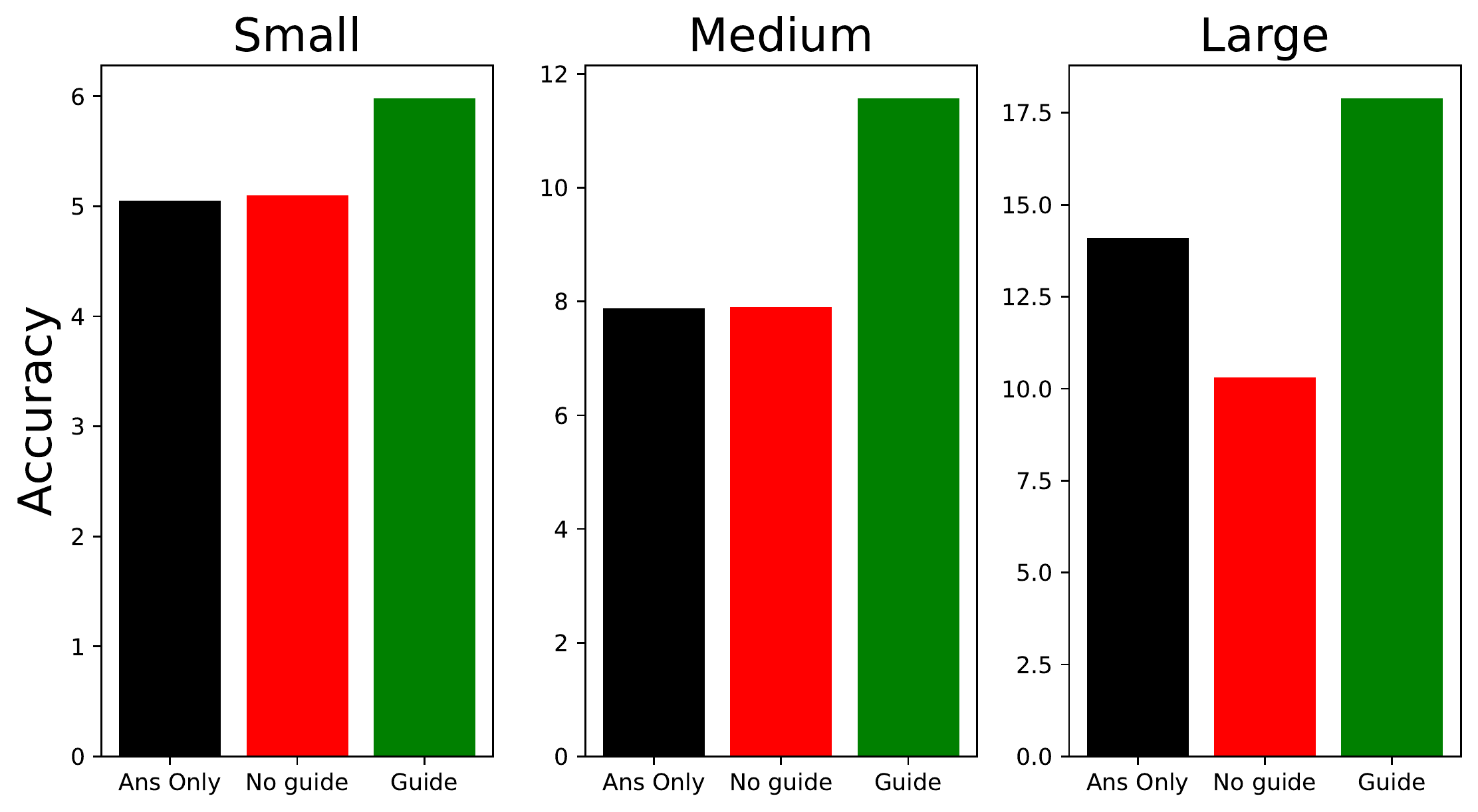}
    \caption{Accuracy of student models (QA + QG) when the question generation is conditioned using the guidance model (Guide) and with non-guided question generation (No guide). Ans only represents the baseline. All models are GPT-2 versions.}
    \label{fig:ablation-guide}
\end{figure}

We found that the fine-tuned GPT-2 predicted an incorrect number of subquestions for the majority of problems (see Table \ref{qg-scores}, first row). Thus, following previous work on subquestion generation \cite{shridhar2022automatic}, we introduced a \textit{guidance mechanism} that conditions the generation of subquestions for a problem $P$ on the equations describing the intermediate solutions of $P$. This strategy improved the quality of the generated questions for all three metrics considered (Table \ref{qg-scores}, second row).
To avoid the dependence on the step-by-step annotation of the equations for each problem $P$ at inference time, we train an additional sequence-to-sequence model to predict, given $P$, the set of equations that lead to the solution of the problem.
At inference time, the predictions for the guidance model are used to condition the generation by the QG model.
Although the predicted equations often do not lead to the correct solution of the problem, they help the QG model to generate more meaningful sub-questions. Figure \ref{fig:ablation-guide} shows the overall accuracy of the GPT-2 student models (QA + QG) fine-tuned with \ourmodel\ on the GSM8K data with and without equation conditioning provided by the guide model.
We have extended this guidance mechanism to StrategyQA and SVAMP, where the generation of subquestions is conditioned on the number of facts (StrategyQA) or steps (SVAMP) needed to answer the problem.

\begin{table} [t]
\centering
\small
\begin{tabular}{c c c c }
    \toprule 
    \bf Methodology  & {\bf BLEU} & \bf BERT $F_1$ & \bf \# Q\\
    \midrule
     No-guidance   & 51.5 & 0.78 & 0.42    \\
    Guidance   & \textbf{58.8} & \bf 0.81 & \bf 0.80    \\
    \bottomrule
\end{tabular}

\caption{BLEU, BERT $F_1$ and the number of questions (\# Q) comparison between the question generator model and the Socratic subquestions present in the GSM8K dataset using GPT2-large model.}
\label{qg-scores}
\end{table}

\paragraph{Eliminating the need for a subquestion module.}
We have experimented with an alternative training solution that does not involve a question-generation model. This strategy aims to improve the supervision for fine-tuning a small model through subquestioning, but without relying on the presence of subquestions at test time. 
The procedure consists of training the student model to generate the entire chain of steps leading to an intermediate answer. That is, when the sub-question $q^{(1)}$ is asked, the model is trained to generate the answer $s^{(1)}$, but when $q^{(j)}$ is asked, the model is trained to generate the chain of thought reasoning $\{s^{(1)}, s^{(2)}, \dots, s^{(j)}\}$ (instead of just $s^{(j)}$). This eliminates the need for the intermediate sub-questions at inference time, as the model is trained to \textit{implicitly} decompose the main problem into smaller reasoning steps. However, this method leads to significant performance degradation (results are reported in Table \ref{Acc:subques-ablation}), highlighting the need for subquestions at inference time.

\begin{figure}[t]
    \centering
    \includegraphics[width=0.47\textwidth]{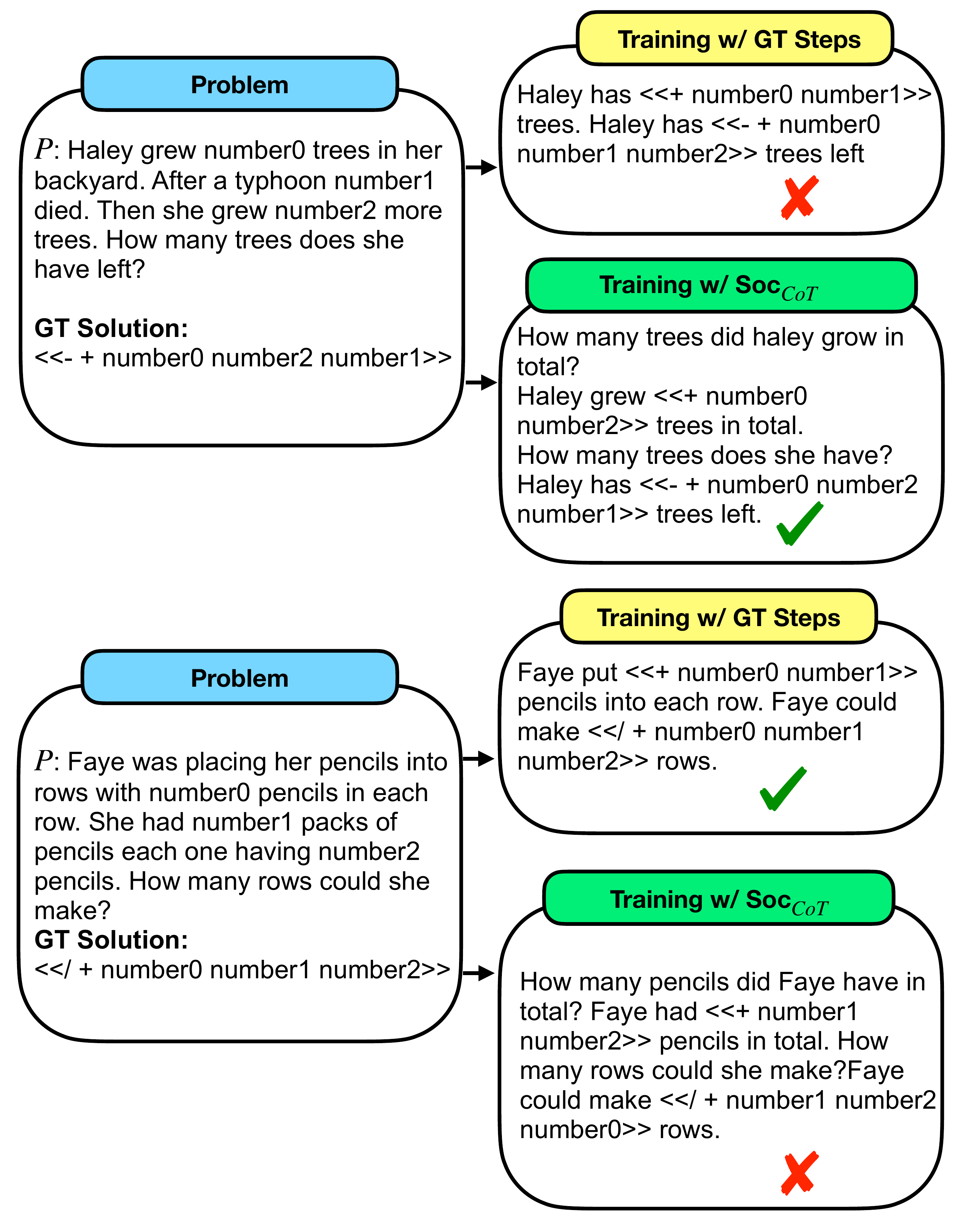}
    \caption{Example of predictions generated by a GPT-2 Medium model fine-tuned with GT steps and \ourmodel\ on the SVAMP dataset.}
    \label{fig:examples_svamp}
\end{figure}

\paragraph{Example outputs}
In Figures \ref{fig:examples} and \ref{fig:examples_svamp}, we report example outputs predicted by GPT-2 models for a set of GSM8K and SVAMP problems.

\begin{table} [t]
\centering
\small
\begin{tabular}{c c c}
    \toprule
    { \bf GPT-2 } & \bf No SubQ & {\bf SubQ with QG } \\
    \midrule
    Small  & 2.70 & \bf 5.98    \\  
    Medium  &  7.20 &  \bf 11.57      \\ 
    Large & 8.18 &  \bf 17.89     \\
    \bottomrule
\end{tabular}
\caption{Accuracy comparison (in \%) of student models trained with (SubQ with QG) and without (No SubQ) question generation model on GSM8K.}
\label{Acc:subques-ablation}
\end{table}

\section{Conclusion }
The chain-of-thought style of step-by-step reasoning has proven to be very effective for reasoning in LLMs. In this work, we propose ways to distill these reasoning capabilities into smaller models and suggest ways to further improve them by explicitly asking stepwise questions. We demonstrate the effectiveness of our proposed methodology on three popular multi-step reasoning datasets, and discuss cases where one method should be preferred over the other for different datasets.

\section*{Limitations}

In our work, we use only one solution from the LLM to distill information into the student model, and according to \citet{Wang2022SelfConsistency}, multiple subquestion-solution pairs can be sampled, and using majority voting, all pairs leading to the most frequent answer can be used to distill knowledge into the student models. Also, due to computational budget, we used a single prompt to compare the CoT and \ourmodel\, and using more prompts (up to 8) might lead to a fairer comparison and better results \cite{wei2022chain}. We leave these experiments for the future. 

\section*{Ethical Considerations}
Although this work improves the reasoning capabilities of smaller models, the models are still not powerful enough to be used in sensitive settings such as education. We plan to release our code and model checkpoints, but the models must be used carefully by users, as many generative models, including ours, are prone to hallucination. 

\section*{Acknowledgements}
Alessandro Stolfo is supported by Armasuisse Science and Technology through a CYD Doctoral Fellowship.

\newpage
\bibliography{custom}

\begin{thebibliography}{42}
\expandafter\ifx\csname natexlab\endcsname\relax\def\natexlab#1{#1}\fi

\bibitem[{Amini et~al.(2019)Amini, Gabriel, Lin, Koncel-Kedziorski, Choi, and
  Hajishirzi}]{amini2019mathqa}
Aida Amini, Saadia Gabriel, Peter Lin, Rik Koncel-Kedziorski, Yejin Choi, and
  Hannaneh Hajishirzi. 2019.
\newblock Mathqa: Towards interpretable math word problem solving with
  operation-based formalisms.
\newblock \emph{arXiv preprint arXiv:1905.13319}.

\bibitem[{Ba and Caruana(2014)}]{ba2014deep}
Jimmy Ba and Rich Caruana. 2014.
\newblock Do deep nets really need to be deep?
\newblock \emph{Advances in neural information processing systems}, 27.

\bibitem[{Brown et~al.(2020)Brown, Mann, Ryder, Subbiah, Kaplan, Dhariwal,
  Neelakantan, Shyam, Sastry, Askell et~al.}]{brown2020language}
Tom Brown, Benjamin Mann, Nick Ryder, Melanie Subbiah, Jared~D Kaplan, Prafulla
  Dhariwal, Arvind Neelakantan, Pranav Shyam, Girish Sastry, Amanda Askell,
  et~al. 2020.
\newblock Language models are few-shot learners.
\newblock \emph{Advances in neural information processing systems},
  33:1877--1901.

\bibitem[{Bruner(1961)}]{bruner1961act}
Jerome~S Bruner. 1961.
\newblock The act of discovery.
\newblock \emph{Harvard educational review}, 31:21--32.

\bibitem[{Cho et~al.(2014)Cho, Van~Merri{\"e}nboer, Gulcehre, Bahdanau,
  Bougares, Schwenk, and Bengio}]{cho2014learning}
Kyunghyun Cho, Bart Van~Merri{\"e}nboer, Caglar Gulcehre, Dzmitry Bahdanau,
  Fethi Bougares, Holger Schwenk, and Yoshua Bengio. 2014.
\newblock Learning phrase representations using rnn encoder-decoder for
  statistical machine translation.
\newblock \emph{arXiv preprint arXiv:1406.1078}.

\bibitem[{Chowdhery et~al.(2022)Chowdhery, Narang, Devlin, Bosma, Mishra,
  Roberts, Barham, Chung, Sutton, Gehrmann et~al.}]{chowdhery2022palm}
Aakanksha Chowdhery, Sharan Narang, Jacob Devlin, Maarten Bosma, Gaurav Mishra,
  Adam Roberts, Paul Barham, Hyung~Won Chung, Charles Sutton, Sebastian
  Gehrmann, et~al. 2022.
\newblock Palm: Scaling language modeling with pathways.
\newblock \emph{arXiv preprint arXiv:2204.02311}.

\bibitem[{Chung et~al.(2022)Chung, Hou, Longpre, Zoph, Tay, Fedus, Li, Wang,
  Dehghani, Brahma et~al.}]{flan-t5}
Hyung~Won Chung, Le~Hou, Shayne Longpre, Barret Zoph, Yi~Tay, William Fedus,
  Eric Li, Xuezhi Wang, Mostafa Dehghani, Siddhartha Brahma, et~al. 2022.
\newblock Scaling instruction-finetuned language models.
\newblock \emph{arXiv preprint arXiv:2210.11416}.

\bibitem[{Cobbe et~al.(2021)Cobbe, Kosaraju, Bavarian, Hilton, Nakano, Hesse,
  and Schulman}]{cobbe2021training}
Karl Cobbe, Vineet Kosaraju, Mohammad Bavarian, Jacob Hilton, Reiichiro Nakano,
  Christopher Hesse, and John Schulman. 2021.
\newblock Training verifiers to solve math word problems.
\newblock \emph{arXiv preprint arXiv:2110.14168}.

\bibitem[{Eisenstein et~al.(2022)Eisenstein, Andor, Bohnet, Collins, and
  Mimno}]{eisenstein2022honest}
Jacob Eisenstein, Daniel Andor, Bernd Bohnet, Michael Collins, and David Mimno.
  2022.
\newblock Honest students from untrusted teachers: Learning an interpretable
  question-answering pipeline from a pretrained language model.
\newblock \emph{arXiv preprint arXiv:2210.02498}.

\bibitem[{Geva et~al.(2021)Geva, Khashabi, Segal, Khot, Roth, and
  Berant}]{geva2021strategyqa}
Mor Geva, Daniel Khashabi, Elad Segal, Tushar Khot, Dan Roth, and Jonathan
  Berant. 2021.
\newblock {Did Aristotle Use a Laptop? A Question Answering Benchmark with
  Implicit Reasoning Strategies}.
\newblock \emph{Transactions of the Association for Computational Linguistics
  (TACL)}.

\bibitem[{Hinton et~al.(2015)Hinton, Vinyals, Dean
  et~al.}]{hinton2015distilling}
Geoffrey Hinton, Oriol Vinyals, Jeff Dean, et~al. 2015.
\newblock Distilling the knowledge in a neural network.
\newblock \emph{arXiv preprint arXiv:1503.02531}, 2(7).

\bibitem[{Hoffmann et~al.(2022)Hoffmann, Borgeaud, Mensch, Buchatskaya, Cai,
  Rutherford, Casas, Hendricks, Welbl, Clark et~al.}]{hoffmann2022training}
Jordan Hoffmann, Sebastian Borgeaud, Arthur Mensch, Elena Buchatskaya, Trevor
  Cai, Eliza Rutherford, Diego de~Las Casas, Lisa~Anne Hendricks, Johannes
  Welbl, Aidan Clark, et~al. 2022.
\newblock Training compute-optimal large language models.
\newblock \emph{arXiv preprint arXiv:2203.15556}.

\bibitem[{Hosseini et~al.(2014)Hosseini, Hajishirzi, Etzioni, and
  Kushman}]{hosseini2014learning}
Mohammad~Javad Hosseini, Hannaneh Hajishirzi, Oren Etzioni, and Nate Kushman.
  2014.
\newblock Learning to solve arithmetic word problems with verb categorization.
\newblock In \emph{Proceedings of the 2014 Conference on Empirical Methods in
  Natural Language Processing (EMNLP)}, pages 523--533.

\bibitem[{Huang et~al.(2022)Huang, Gu, Hou, Wu, Wang, Yu, and
  Han}]{huang2022large}
Jiaxin Huang, Shixiang~Shane Gu, Le~Hou, Yuexin Wu, Xuezhi Wang, Hongkun Yu,
  and Jiawei Han. 2022.
\newblock Large language models can self-improve.
\newblock \emph{arXiv preprint arXiv:2210.11610}.

\bibitem[{Iyer et~al.(2022)Iyer, Lin, Pasunuru, Mihaylov, Simig, Yu, Shuster,
  Wang, Liu, Koura et~al.}]{iyer2022opt}
Srinivasan Iyer, Xi~Victoria Lin, Ramakanth Pasunuru, Todor Mihaylov,
  D{\'a}niel Simig, Ping Yu, Kurt Shuster, Tianlu Wang, Qing Liu, Punit~Singh
  Koura, et~al. 2022.
\newblock Opt-iml: Scaling language model instruction meta learning through the
  lens of generalization.
\newblock \emph{arXiv preprint arXiv:2212.12017}.

\bibitem[{Klein and Nabi(2019)}]{klein2019learning}
Tassilo Klein and Moin Nabi. 2019.
\newblock Learning to answer by learning to ask: Getting the best of gpt-2 and
  bert worlds.
\newblock \emph{arXiv preprint arXiv:1911.02365}.

\bibitem[{Kojima et~al.(2022)Kojima, Gu, Reid, Matsuo, and
  Iwasawa}]{kojima2022large}
Takeshi Kojima, Shixiang~Shane Gu, Machel Reid, Yutaka Matsuo, and Yusuke
  Iwasawa. 2022.
\newblock Large language models are zero-shot reasoners.
\newblock \emph{arXiv preprint arXiv:2205.11916}.

\bibitem[{Kushman et~al.(2014)Kushman, Artzi, Zettlemoyer, and
  Barzilay}]{kushman-etal-2014-learning}
Nate Kushman, Yoav Artzi, Luke Zettlemoyer, and Regina Barzilay. 2014.
\newblock \href {https://doi.org/10.3115/v1/P14-1026} {Learning to
  automatically solve algebra word problems}.
\newblock In \emph{Proceedings of the 52nd Annual Meeting of the Association
  for Computational Linguistics (Volume 1: Long Papers)}, pages 271--281,
  Baltimore, Maryland. Association for Computational Linguistics.

\bibitem[{Lewkowycz et~al.(2022)Lewkowycz, Andreassen, Dohan, Dyer,
  Michalewski, Ramasesh, Slone, Anil, Schlag, Gutman-Solo
  et~al.}]{lewkowycz2022solving}
Aitor Lewkowycz, Anders Andreassen, David Dohan, Ethan Dyer, Henryk
  Michalewski, Vinay Ramasesh, Ambrose Slone, Cem Anil, Imanol Schlag, Theo
  Gutman-Solo, et~al. 2022.
\newblock Solving quantitative reasoning problems with language models.
\newblock \emph{arXiv preprint arXiv:2206.14858}.

\bibitem[{Li et~al.(2022)Li, Chen, Shen, Chen, Zhang, Li, Wang, Qian, Peng, Mao
  et~al.}]{li2022explanations}
Shiyang Li, Jianshu Chen, Yelong Shen, Zhiyu Chen, Xinlu Zhang, Zekun Li, Hong
  Wang, Jing Qian, Baolin Peng, Yi~Mao, et~al. 2022.
\newblock Explanations from large language models make small reasoners better.
\newblock \emph{arXiv preprint arXiv:2210.06726}.

\bibitem[{Miao et~al.(2020)Miao, Liang, and Su}]{miao-etal-2020-diverse}
Shen-yun Miao, Chao-Chun Liang, and Keh-Yih Su. 2020.
\newblock \href {https://doi.org/10.18653/v1/2020.acl-main.92} {A diverse
  corpus for evaluating and developing {E}nglish math word problem solvers}.
\newblock In \emph{Proceedings of the 58th Annual Meeting of the Association
  for Computational Linguistics}, pages 975--984, Online. Association for
  Computational Linguistics.

\bibitem[{Nye et~al.(2021)Nye, Andreassen, Gur-Ari, Michalewski, Austin,
  Bieber, Dohan, Lewkowycz, Bosma, Luan et~al.}]{nye2021show}
Maxwell Nye, Anders~Johan Andreassen, Guy Gur-Ari, Henryk Michalewski, Jacob
  Austin, David Bieber, David Dohan, Aitor Lewkowycz, Maarten Bosma, David
  Luan, et~al. 2021.
\newblock Show your work: Scratchpads for intermediate computation with
  language models.
\newblock \emph{arXiv preprint arXiv:2112.00114}.

\bibitem[{Opedal et~al.(2023)Opedal, Stoehr, Saparov, and
  Sachan}]{opedal-et-al-2023}
Andreas Opedal, Niklas Stoehr, Abulhair Saparov, and Mrinmaya Sachan. 2023.
\newblock World models for math story problems.
\newblock In \emph{Findings of the Association for Computational Linguistics:
  ACL 2023}, Toronto, Canada.

\bibitem[{Patel et~al.(2021)Patel, Bhattamishra, and Goyal}]{patel2021nlp}
Arkil Patel, Satwik Bhattamishra, and Navin Goyal. 2021.
\newblock Are nlp models really able to solve simple math word problems?
\newblock \emph{arXiv preprint arXiv:2103.07191}.

\bibitem[{Post(2018)}]{sacre-bleu}
Matt Post. 2018.
\newblock \href {https://www.aclweb.org/anthology/W18-6319} {A call for clarity
  in reporting {BLEU} scores}.
\newblock In \emph{Proceedings of the Third Conference on Machine Translation:
  Research Papers}, pages 186--191, Belgium, Brussels. Association for
  Computational Linguistics.

\bibitem[{Radford et~al.(2019)Radford, Wu, Child, Luan, Amodei, and
  Sutskever}]{radford2019language}
Alec Radford, Jeff Wu, Rewon Child, David Luan, Dario Amodei, and Ilya
  Sutskever. 2019.
\newblock Language models are unsupervised multitask learners.

\bibitem[{Rajani et~al.(2019)Rajani, McCann, Xiong, and
  Socher}]{rajani_explain_2019}
Nazneen~Fatema Rajani, Bryan McCann, Caiming Xiong, and Richard Socher. 2019.
\newblock \href {https://doi.org/10.18653/v1/P19-1487} {Explain {Yourself}!
  {Leveraging} {Language} {Models} for {Commonsense} {Reasoning}}.
\newblock In \emph{Proceedings of the 57th {Annual} {Meeting} of the
  {Association} for {Computational} {Linguistics}}, pages 4932--4942, Florence,
  Italy. Association for Computational Linguistics.

\bibitem[{Rao and Daumé~III()}]{rao_answer-based_nodate}
Sudha Rao and Hal Daumé~III.
\newblock \href {https://www.aclweb.org/anthology/N19-1013/} {Answer-based
  {Adversarial} {Training} for {Generating} {Clarification} {Questions}}.
\newblock In \emph{Proceedings of the 2019 {Conference} of the {North}
  {American} {Chapter} of the {Association} for {Computational} {Linguistics}:
  {Human} {Language} {Technologies}, {Volume} 1 ({Long} and {Short} {Papers})}.

\bibitem[{Roy et~al.(2015)Roy, Vieira, and Roth}]{roy-etal-2015-reasoning}
Subhro Roy, Tim Vieira, and Dan Roth. 2015.
\newblock \href {https://doi.org/10.1162/tacl_a_00118} {Reasoning about
  quantities in natural language}.
\newblock \emph{Transactions of the Association for Computational Linguistics},
  3:1--13.

\bibitem[{Shridhar et~al.(2022)Shridhar, Macina, El-Assady, Sinha, Kapur, and
  Sachan}]{shridhar2022automatic}
Kumar Shridhar, Jakub Macina, Mennatallah El-Assady, Tanmay Sinha, Manu Kapur,
  and Mrinmaya Sachan. 2022.
\newblock Automatic generation of socratic subquestions for teaching math word
  problems.
\newblock \emph{arXiv preprint arXiv:2211.12835}.

\bibitem[{Shwartz et~al.(2020)Shwartz, West, Le~Bras, Bhagavatula, and
  Choi}]{shwartz-etal-2020-unsupervised}
Vered Shwartz, Peter West, Ronan Le~Bras, Chandra Bhagavatula, and Yejin Choi.
  2020.
\newblock \href {https://doi.org/10.18653/v1/2020.emnlp-main.373} {Unsupervised
  commonsense question answering with self-talk}.
\newblock In \emph{Proceedings of the 2020 Conference on Empirical Methods in
  Natural Language Processing (EMNLP)}, pages 4615--4629, Online. Association
  for Computational Linguistics.

\bibitem[{Snell et~al.(2022)Snell, Klein, and Zhong}]{snell2022learning}
Charlie Snell, Dan Klein, and Ruiqi Zhong. 2022.
\newblock Learning by distilling context.
\newblock \emph{arXiv preprint arXiv:2209.15189}.

\bibitem[{Srivastava et~al.(2022)Srivastava, Rastogi, Rao, Shoeb, Abid, Fisch,
  Brown, Santoro, Gupta, Garriga-Alonso et~al.}]{srivastava2022beyond}
Aarohi Srivastava, Abhinav Rastogi, Abhishek Rao, Abu Awal~Md Shoeb, Abubakar
  Abid, Adam Fisch, Adam~R Brown, Adam Santoro, Aditya Gupta, Adri{\`a}
  Garriga-Alonso, et~al. 2022.
\newblock Beyond the imitation game: Quantifying and extrapolating the
  capabilities of language models.
\newblock \emph{arXiv preprint arXiv:2206.04615}.

\bibitem[{Stolfo et~al.(2022)Stolfo, Jin, Shridhar, Sch{\"o}lkopf, and
  Sachan}]{stolfo2022causal}
Alessandro Stolfo, Zhijing Jin, Kumar Shridhar, Bernhard Sch{\"o}lkopf, and
  Mrinmaya Sachan. 2022.
\newblock A causal framework to quantify the robustness of mathematical
  reasoning with language models.
\newblock \emph{arXiv preprint arXiv:2210.12023}.

\bibitem[{Vaswani et~al.(2017)Vaswani, Shazeer, Parmar, Uszkoreit, Jones,
  Gomez, Kaiser, and Polosukhin}]{vaswani2017attention}
Ashish Vaswani, Noam Shazeer, Niki Parmar, Jakob Uszkoreit, Llion Jones,
  Aidan~N Gomez, {\L}ukasz Kaiser, and Illia Polosukhin. 2017.
\newblock Attention is all you need.
\newblock \emph{Advances in neural information processing systems}, 30.

\bibitem[{Wang et~al.(2022)Wang, Wei, Schuurmans, Le, Chi, and
  Zhou}]{Wang2022SelfConsistency}
Xuezhi Wang, Jason Wei, Dale Schuurmans, Quoc Le, Ed~Chi, and Denny Zhou. 2022.
\newblock Self-consistency improves chain of thought reasoning in language
  models.
\newblock \emph{ArXiv}, abs/2203.11171.

\bibitem[{Wei et~al.(2022{\natexlab{a}})Wei, Tay, Bommasani, Raffel, Zoph,
  Borgeaud, Yogatama, Bosma, Zhou, Metzler et~al.}]{wei2022emergent}
Jason Wei, Yi~Tay, Rishi Bommasani, Colin Raffel, Barret Zoph, Sebastian
  Borgeaud, Dani Yogatama, Maarten Bosma, Denny Zhou, Donald Metzler, et~al.
  2022{\natexlab{a}}.
\newblock Emergent abilities of large language models.
\newblock \emph{arXiv preprint arXiv:2206.07682}.

\bibitem[{Wei et~al.(2022{\natexlab{b}})Wei, Wang, Schuurmans, Bosma, Chi, Le,
  and Zhou}]{wei2022chain}
Jason Wei, Xuezhi Wang, Dale Schuurmans, Maarten Bosma, Ed~Chi, Quoc Le, and
  Denny Zhou. 2022{\natexlab{b}}.
\newblock Chain of thought prompting elicits reasoning in large language
  models.
\newblock \emph{arXiv preprint arXiv:2201.11903}.

\bibitem[{Wolf et~al.(2020)Wolf, Debut, Sanh, Chaumond, Delangue, Moi, Cistac,
  Rault, Louf, Funtowicz, Davison, Shleifer, von Platen, Ma, Jernite, Plu, Xu,
  Le~Scao, Gugger, Drame, Lhoest, and Rush}]{wolf2019huggingface}
Thomas Wolf, Lysandre Debut, Victor Sanh, Julien Chaumond, Clement Delangue,
  Anthony Moi, Pierric Cistac, Tim Rault, Remi Louf, Morgan Funtowicz, Joe
  Davison, Sam Shleifer, Patrick von Platen, Clara Ma, Yacine Jernite, Julien
  Plu, Canwen Xu, Teven Le~Scao, Sylvain Gugger, Mariama Drame, Quentin Lhoest,
  and Alexander Rush. 2020.
\newblock \href {https://doi.org/10.18653/v1/2020.emnlp-demos.6} {Transformers:
  State-of-the-art natural language processing}.
\newblock In \emph{Proceedings of the 2020 Conference on Empirical Methods in
  Natural Language Processing: System Demonstrations}, pages 38--45, Online.
  Association for Computational Linguistics.

\bibitem[{Zhang et~al.(2020)Zhang, Wang, Lee, Bin, Wang, Shao, and
  Lim}]{zhang2020graph}
Jipeng Zhang, Lei Wang, Roy Ka-Wei Lee, Yi~Bin, Yan Wang, Jie Shao, and Ee-Peng
  Lim. 2020.
\newblock Graph-to-tree learning for solving math word problems.
\newblock Association for Computational Linguistics.

\bibitem[{Zhang et~al.(2019)Zhang, Kishore, Wu, Weinberger, and
  Artzi}]{zhang2019bertscore}
Tianyi Zhang, Varsha Kishore, Felix Wu, Kilian~Q Weinberger, and Yoav Artzi.
  2019.
\newblock Bertscore: Evaluating text generation with bert.
\newblock \emph{arXiv preprint arXiv:1904.09675}.

\bibitem[{Zhou et~al.(2022)Zhou, Scharli, Hou, Wei, Scales, Wang, Schuurmans,
  Bousquet, Le, and Chi}]{Zhou2022LeasttoMostPE}
Denny Zhou, Nathanael Scharli, Le~Hou, Jason Wei, Nathan Scales, Xuezhi Wang,
  Dale Schuurmans, Olivier Bousquet, Quoc Le, and Ed~Chi. 2022.
\newblock Least-to-most prompting enables complex reasoning in large language
  models.
\newblock \emph{ArXiv}, abs/2205.10625.

\end{thebibliography}
\bibliographystyle{acl_natbib}

\newpage
\appendix

\begin{table*}[]\small
\resizebox{\textwidth}{!}{
    \centering
    \begin{tabularx}{\textwidth}{ X }
    \toprule[0.1em] \\
        Let's generate sub-questions for these problems. Use exactly one operation per step. \\
\\
---\\
Q: Zoe was unboxing some of her old winter clothes . She found number0 boxes of clothing and inside each box there were number1 scarves and number2 mittens . How many pieces of winter clothing did Zoe have total ?\\
\\
SQ1: How many pieces of winter clothing did Zoe have in each box?\\
A1: Zoe had \texttt{<<}+ number1 number2\texttt{>>} pieces of winter clothing in each box.\\
\\
SQ2: How many pieces of winter clothing did Zoe have total ?\\
A2: Zoe had \texttt{<<}* number0 + number1 number2\texttt{>>} pieces of winter clothing in total.\\
\\
---\\
Q: Katie picked number0 tulips and number1 roses to make flower bouquets . If she only used number2 of the flowers though , how many extra flowers did Katie pick ?\\
\\
SQ1: How many flowers did Katie pick in total?\\
A1: Katie picked \texttt{<<}+ number0 number1\texttt{>>} flowers in total.\\
\\
SQ2: How many extra flowers did Katie pick ?\\
A2: Katie picked \texttt{<<}- + number0 number1 number2\texttt{>>} extra flowers.\\
\\
---\\
Q: Conner has number0 dollars in his bank account . Every month he spends number1 dollars . He does not add money to the account . How much money will Conner have in his account after number2 months ?,\\
\\
SQ1: How much money does Conner spend in total?
A1: Conner spends \texttt{<<}* number1 number2\texttt{>>} dollars.

SQ2: How much money will Conner have in his account after 8.0 months ?
A2: After 8.0 months, Conner will have <<- number0 * number1 number2\texttt{>>} dollars.\\
\midrule
For each of the following topics, generate intermediate answers to the subquestions leading to the final answer.\\
\\
---\\
Topic: Albany, Georgia (City in Georgia, United States)\\
Will the Albany in Georgia reach a hundred thousand occupants before the one in New York?\\
\\
Albany, GA has around 75,000 people.\\
Albany, NY has almost 100,000 people.\\
The difference is 100,000-75,000=25,000\\
The difference is 100,000-100,000=0\\
No, 25,000 is not smaller than 0.\\
The final answer is NO.\\
\\
---\\
Topic: The Police (English rock band)\\
Could the members of The Police perform lawful arrests?\\
\\
Only law enforcement officers can perform lawful arrests.\\
No, the members of The Police (rock band) are not law enforcement officers.\\
The final answer is NO.\\
\\
---\\
Topic: Wonder Woman (2017 film) (American superhero film directed by Patty Jenkins)
Is a Boeing 737 cost covered by Wonder Woman (2017 film) box office receipts?\\
\\
The average cost of a US Boeing 737 plane is 1.6 million dollars.\\
Wonder Woman (2017 film) grossed over 800 million dollars at the box office.\\
Yes, 800 is larger than 1.6.\\
The final answer is YES.\\

\bottomrule[0.1em]
    \end{tabularx}
    }
    \caption{Exemplars included in the few-shot prompt for the decomposition of the problems from the ASDiv (upper row) and StrategyQA (lower row) datasets.}
    \label{table:prompts}
\end{table*}

\end{document}